\documentclass[runningheads]{llncs}
\usepackage[T1]{fontenc}
\usepackage{graphicx,verbatim}
\usepackage{amssymb}
\usepackage{adjustbox}
\usepackage{amsmath}
\usepackage{multirow}
\usepackage{booktabs}
\usepackage{tikz}
\usepackage{wasysym}
\usepackage{xcolor}
\usepackage{tabularx}
\usepackage[colorlinks=true,
            linkcolor=blue,
            anchorcolor=blue,
            citecolor=blue,
            urlcolor=blue,
            ]{hyperref}
\usepackage{pifont}  % 在导言区添加
\newcolumntype{Y}{>{\centering\arraybackslash}X}
\usepackage{makecell}  % 使用 makecell 来强制居中
\title{Bridging the Gap in Missing Modalities: Leveraging Knowledge Distillation and Style Matching for Brain Tumor Segmentation}
\author{%
  Shenghao Zhu\inst{1}\thanks{Co‑first author.} \and
  Yifei Chen\inst{2}\protect\footnotemark[1] \and
  Weihong Chen\inst{1} \and
  Yuanhan Wang\inst{1} \and
  Chang Liu\inst{1} \and
  Shuo Jiang\inst{1} \and
  Feiwei Qin\inst{1}\thanks{Corresponding author.} \and
  Changmiao Wang\inst{3}\protect\footnotemark[2]
}
\authorrunning{S. Zhu et al.}
\titlerunning{Bridging the Gap in Missing Modalities}
\institute{%
  Hangzhou Dianzi University, Hangzhou, China  \and
  Tsinghua University, Beijing, China          \and
  Shenzhen Research Institute of Big Data, Shenzhen, China\\
  \email{qinfeiwei@hdu.edu.cn, cmwangalbert@gmail.com}
}

\begin{document}
\maketitle              % typeset the header of the contribution
\begin{abstract}
Accurate and reliable brain tumor segmentation, particularly when dealing with missing modalities, remains a critical challenge in medical image analysis. Previous studies have not fully resolved the challenges of tumor boundary segmentation insensitivity and feature transfer in the absence of key imaging modalities. In this study, we introduce MST-KDNet, aimed at addressing these critical issues. Our model features Multi-Scale Transformer Knowledge Distillation to effectively capture attention weights at various resolutions, Dual-Mode Logit Distillation to improve the transfer of knowledge, and a Global Style Matching Module that integrates feature matching with adversarial learning. Comprehensive experiments conducted on the BraTS and FeTS 2024 datasets demonstrate that MST-KDNet surpasses current leading methods in both Dice and HD95 scores, particularly in conditions with substantial modality loss. Our approach shows exceptional robustness and generalization potential, making it a promising candidate for real-world clinical applications. Our source code is available at \href{https://github.com/Quanato607/MST-KDNet}{https://github.com/Quanato607/MST-KDNet}.

% Authors must provide keywords and are not allowed to remove this Keyword section.

\end{abstract}
\section{Introduction}

Brain tumor segmentation is a critical task in medical neuroimaging, playing a vital role in diagnosis, treatment planning, and prognosis assessment. Among brain tumors, gliomas stand out as particularly aggressive, exhibiting complex biological behaviors and significant heterogeneity, which complicate clinical treatment \cite{c1weller2015glioma}. Magnetic resonance imaging (MRI), utilizing multiple modalities, is the preferred method for visualizing and segmenting brain tumors due to its ability to capture diverse and complementary information \cite{c2zhu2024brain,c27wu2025towards,c28chen2025toward}. Each MRI modality contributes unique insights: T1 and T2 modalities are effective in identifying angioedema in subacute strokes; T1Gd highlights vascular structures and the blood-brain barrier; and FLAIR provides a broad overview of stroke lesion characteristics. Together, these modalities complement one another, offering detailed information about tumor size, location, and morphology \cite{c18zhu2024xlstm}.

Missing modalities, caused by scan corruption, imaging artifacts, and varying machine settings, are a common challenge in clinical settings \cite{c19zhang2024tc,c24liu2024anomaly,c26ge2024tc}. Furthermore, in clinical practice, the number of available modalities and their correct labeling for algorithmic use is uncertain, limiting segmentation accuracy \cite{c3varsavsky2018pimms}. Several methods have been proposed to handle missing modalities, such as Ding \textit{et al.}'s \cite{c4ding2021rfnet} region-aware fusion module and Zhao \textit{et al.}'s \cite{c5zhao2022modality} modality-adaptive feature interaction. Zhang \textit{et al.} \cite{c6zhang2023multi} and Dai \textit{et al.} \cite{c7dai2024federated} introduced techniques to reconstruct missing modalities by aggregating features from available ones. Whereas, the absence of critical modalities often leads to performance degradation \cite{c8wang2023learnable}. To tackle this, Wang \textit{et al.} \cite{c8wang2023learnable} proposed a cross-modal knowledge distillation framework to identify key modalities. In contrast, Liu \textit{et al.} \cite{c9liu2023m3ae} developed M3AE using self-distillation to handle missing-modality. Other approaches, such as Huo \textit{et al.}'s \cite{c10huo2024c2kd} bi-directional distillation and Xing \textit{et al.}'s \cite{c11xing2024comprehensive} contrastive distillation, have proved the efficacy of knowledge distillation in improving model performance despite missing modalities. However, past methods remain inadequate in handling modality inconsistency and correlation due to limited knowledge adaptability and the inability to align features between modalities.

To address the difficulty of understanding the inter-modal correlation and inconsistency in the model, this paper proposes \textbf{MST-KDNet}, which significantly improves cross-modal semantic extraction, maintaining superior segmentation accuracy and robust generalization capabilities across varying modality combinations. The main contributions of this work are summarized as follows: 1) We propose the Multi-Scale Transformer Knowledge Distillation (MS-TKD), which enhances segmentation performance by extracting attention weights and features across different resolutions; 2) We present the Dual-Modal Logit Distillation (DMLD), which leverages Logit Alignment and Normalized Kullback-Leibler Distillation to enhance knowledge adaptation and ensure robust learning even in the absence of certain modalities; 3) We develop the Global Style Matching Module (GSME) to combine feature matching with adversarial learning to strengthen the model’s performance across missing modality scenarios.

\begin{figure}[h!]
    \centering
    \includegraphics[width=1\textwidth]{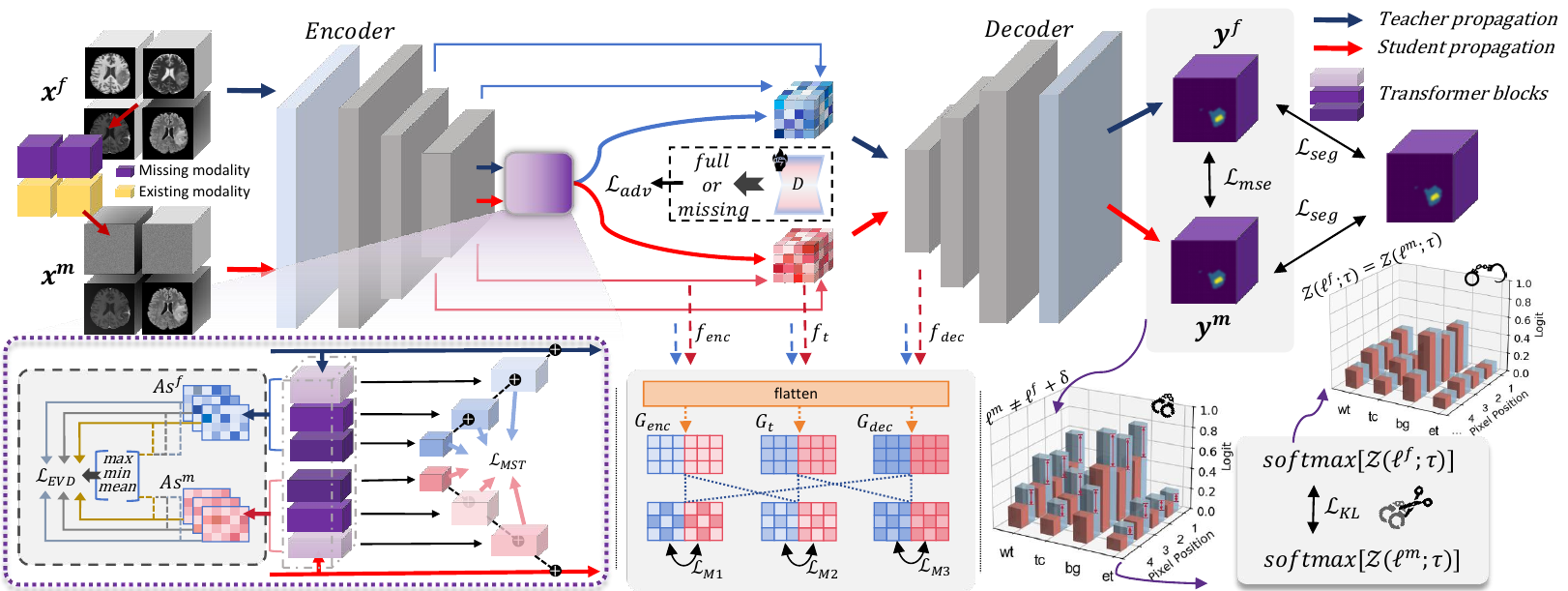}
    \caption{The overall framework of MST-KDNet. The teacher propagation processes all available modalities, while the student propagation accommodates incomplete inputs.}
    \label{fig:1}
\end{figure}

% This model enhances segmentation performance by employing Multi-Scale Transformer Knowledge Distillation (MS-TKD) to extract attention weights across different resolutions. Additionally, it incorporates an advanced extreme value distillation mechanism, enabling the student model to more effectively capture key features from the teacher model. To further improve knowledge transfer, we introduce the Dual-Modal Logit Distillation (DMLD) technique, which leverages Logit Alignment and Normalized KL Distillation to enhance knowledge adaptation. Moreover, the innovative Global Style Matching Module (GSME) combines feature matching with adversarial learning, strengthening the model's ability to perform effectively even when certain modalities are absent.

\section{Proposed Method}
\subsection{Baseline Network}

The baseline network structure of the model is based on 3D convolution and multi-scale Transformer architecture \cite{c20hatamizadeh2022unetr}. First, the input \( x \in \mathbb{R}^{H \times W \times D \times C} \) undergoes one layer of 3D convolution to extract the initial features and then reduces the feature map size layer-by-layer by three layers of 3D convolutional downsampling with a step size of 2, to extract the rich spatial and semantic information in the external encoder stage. After downsampling, the feature map size is \( x \in \mathbb{R}^{(H/8) \times (W/8) \times (D/8) \times 32C} \) and is converted to a 1D sequence. The volume is divided into non-overlapping \( (P, P, P) \) chunks to obtain \( x_v \in \mathbb{R}^{(N \times (P^3 \cdot 32C))} \), with a sequence length of \( N = \frac{(H/8) \cdot (W/8) \cdot (D/8)}{P^3} \). These chunks are projected through a linear layer into the \( K \) dimensional embedding space and enter the Transformer processing. The Transformer architecture contains multiple Transformer blocks using the Multihead Self-Attention (MSA) mechanism. Each MSA sublayer has \( n \) parallel self-attention heads, and the attention weight \( A \) is computed by querying (\( q \)) and key (\( k \)) similarity with the following formula:
\begin{equation}
    A = \operatorname{softmax}\!\left(\frac{q\, k^\top}{\sqrt{K_h}}\right);\text{SA}(z) = A\; v;    \text{MSA}(\mathbf{z}) = 
    \left[ \text{SA}_1(\mathbf{z});...;\text{SA}_n(\mathbf{z}) \right] W,
\end{equation}
where \( K_h = \frac{K}{n} \) serves as a scaling factor to keep the number of parameters consistent across key dimensions. Here, \( v \) denotes the value mapping in the sequence \( z \), and \( W \in \mathbb{R}^{(n \cdot K_h) \times K} \) denotes the trainable parameter weights of the multi-head self-attention sublayer. At different resolution stages, multiple \( z_i \) representations are extracted, sized as \( \frac{H \times W \times D}{P^3} \times K \) and reshaped as \( \frac{H}{P} \times \frac{W}{P} \times \frac{D}{P} \times K \) tensor after \( 3 \times 3 \times 3 \) convolution and normalization. Thereafter, the feature maps are combined with the feature maps of the corresponding Transformer blocks through jump connections. The inverse convolution extends the size. Finally, the external decoder extracts the deep semantic information through 3D convolution and integrates the multi-scale jump-connected features, and up-sampling gradually restores the spatial resolution. 

\subsection{Multi-scale Transformer Knowledge Distillation}
As illustrated in Fig.~\ref{fig:1}, we extract the attention weights \( A \) from each resolution layer as a sequence for the Extreme Value Distillation (EVD) process. We calculate the maximum, minimum, and mean values of these attention weights at each pixel position along the C dimension:
\begin{gather}
    A_{\max} = \max(As, \text{axis} = 1) ;
    A_{\min} = \min(As, \text{axis} = 1) ;
    A_{\text{mean}} = \frac{1}{n} \sum_{i=1}^{n} A_i.
\end{gather}
These values are then used as weights to generate three sequences by multiplying them with the corresponding attention weights:
\begin{gather}
    EV_{1} = A_{\max} \cdot A;
    EV_{2} = A_{\min} \cdot A ;
    EV_{3} = A_{\text{mean}} \cdot A.
\end{gather}
For the complete modality teacher model and the student model with missing modalities, we apply knowledge distillation using mean square error (MSE) loss. Additionally, MSE is applied to each reshaped tensor, ensuring consistency between the models. The multiscale Transformer-based knowledge distillation loss is defined as follows:
\begin{align}
    \mathcal{L}_{\text{MS-TKD}} &= \alpha \mathcal{L}_{\text{EVD}} (EV^f, EV^m) + \beta \mathcal{L}_{\text{MST}}   
  \notag \\
     &= \alpha \frac{1}{n} \sum_{i=1}^{3} \sum_{j=1}^{n}    \Big( EV_{i,j}^{f} - EV_{i,j}^{m} \Big)^2+ \beta \frac{1}{N} \sum_{i=1}^{3} \sum_{j=1}^{N}  (z_{i,j}^f - z_{i,j}^m)^2.
\end{align}
% \begin{align}
%     L_{\text{EVD}} (EV^f, EV^m) &= \frac{1}{2} \sum_{j=1}^{n}  
%     \Big( EV_{\max}^{f} - EV_{\max}^{m} \Big)^2
%     + \Big( EV_{\min}^{f} - EV_{\min}^{m} \Big)^2 \nonumber \\
%     &\quad + \Big( EV_{\text{mean}}^{f} - EV_{\text{mean}}^{m} \Big)^2
% \end{align}
% \begin{align}
%     L_{\text{MS-TKD}} &= \alpha L_{\text{EVD}} (EV^f, EV^m) + \beta L_{\text{MST}} = \frac{1}{2} \sum_{i=1}^{3} \sum_{j}  
%     \Big( EV_{i,j}^{f} - EV_{i,j}^{m} \Big)^2 + \notag \\
%      &\Big( EV_{i,j}^{f} - EV_{i,j}^{m} \Big)^2 + \Big( EV_{\text{mean}}^{f} - EV_{\text{mean}}^{m} \Big)^2 + \beta \frac{1}{2} \sum_{i=1}^{n} \sum_{j}  (z_{i,j}^f - z_{i,j}^m)^2.
% \end{align}

% At the same time, we performed knowledge distillation using MSE for each after reshaping the tensor shape.
% Thus, the loss (MS-TKD) of knowledge distillation based on a multiscale Transformer is:
% \begin{equation}
%     L_{\text{MS-TKD}} = \alpha L_{\text{EVD}} + \beta \frac{1}{2} \sum_{j=1}^{n} (z_j^f - z_j^m)^2
% \end{equation}

\subsection{Dual-mode Logit Distillation}
\textbf{Logit Discrepancy Distillation.} 
To enhance the student's ability to mimic the teacher model's style, we apply MSE loss to align features from the complete modality with those of the missing modality. The 
specific formula for this calculation is as follows:
% To enhance the student's ability to mimic the teacher model's style, we apply MSE loss to align features from the complete modality with those of the missing modality. By computing the mean square error between the logit outputs of the teacher and student models, we enable the features from the missing modality to gradually converge with those of the complete modality. The formula for this calculation is as follows:
\begin{equation}
    \mathcal{L}_{\text{mse}} (l^f, l^m) = \frac{1}{n} \sum_{j=1}^{n} \left( l_j^f - l_j^m \right)^2,
\end{equation}
where \( l^f \) and \( l^m \) represent the logit outputs from the teacher and student models, respectively, with \( N \) indicating the dimension of the logit vector.

\textbf{Logit Standardization KL Distillation.}
Traditional knowledge distillation applies a global temperature factor to align the logit ranges of both student and teacher networks. This rigid coupling limits the student's ability to adapt, especially when there is a significant disparity in model capacity, reducing the student's learning potential. To address this limitation, we introduce logit normalization into the distillation process \cite{c25sun2024logit}. Specifically, the logits (\( l \)) are first normalized using the \(\mathcal{Z}\)-score normalization function before being passed through the softmax function:
\begin{gather}
    \mathcal{Z}(l; \tau) = \frac{1 - \mu}{(\sigma + 10^{-7}) \tau};
    q(l) = \text{softmax}[\mathcal{Z}(l; \tau)] \quad \text{where} \quad l \in \{l^f, l^m\},
\end{gather}
where \(\mu\) represents the mean, \(\sigma\) denotes the standard deviation, and \(\tau\) is the temperature coefficient. The normalized logits are then passed into the KL divergence loss function, defined as:
\begin{equation}
    \mathcal{L}_{\text{KL}} (l^f || l^m) = \sum_{k=1}^{K} q(l^f) \log \left( \frac{q(l^f)}{q(l^m)} \right).
\end{equation}
% Logit Standardization KL Distillation offers several advantages. First, it eliminates the reliance on a globally shared temperature, allowing more flexible adjustments as needed. Second, while the standardization process preserves the core distribution relationship between the teacher and student's logits, it avoids rigidly imposing the teacher's numerical output magnitude on the student. Thus, the Dual-Mode Logit Distillation loss function combines Adversarial Learning loss and Logit Standardization KL loss, expressed as:
Logit Standardization KL Distillation removes the need for a globally shared temperature, allowing for flexible adjustments, and preserves the core distribution relationship between the teacher's and student's logits without rigidly matching the teacher's output magnitude. Thus, the Dual-Mode Logit Distillation loss function combines Logit Discrepancy loss and Logit Standardization KL loss, expressed as:
\begin{equation}
    \mathcal{L}_{\text{logit}} = \lambda_{\text{mse}} \mathcal{L}_{\text{mse}} + \lambda_{\text{KD}} \tau^2 \mathcal{L}_{\text{KL}}.
\end{equation}

\subsection{Global Style Matching Module}
The structural and stylistic variations inherent to different MRI modalities often pose challenges for decoding networks, particularly when some modalities are missing. To address this issue, our GSME integrates Mean Square Error MSE loss with adversarial learning to mitigate these challenges. In GSME, we first take the max-pooled feature output from the penultimate convolutional layer, \(f_{\text{enc}}\), concatenate it with the output of the transformer block, \(f_{\text{t}}\), and feed the combined features into the decoder. The decoder processes these features and outputs \(f_{\text{dec}}\), while the fused features, \(f_{\text{enc} \& \text{t}}\), are simultaneously passed into a feature discriminator \(D\) to compute the adversarial loss:
\begin{equation}
    \mathcal{L}_{\text{adv}} = \log(1 - D(f_{\text{enc} \& \text{t}}^f)) + \log(D(f_{\text{enc} \& \text{t}}^m)),
\end{equation}
where \(f_{\text{enc} \& \text{t}}\) represents the input to the decoder. The decoder's first convolutional output, \(f_{\text{dec}}\), is further processed by reshaping the spatial dimensions \(H\), \(W\), and \(D\) of \(f_{\text{enc}}, f_{\text{t}},\) and \(f_{\text{dec}}\) into two-dimensional tensors \(G_{\text{enc}}, G_{\text{t}}, G_{\text{dec}}\). These tensors are then subjected to a feature fusion operation:
\begin{gather}
    M_{1} = G_{\text{enc}} G_{\text{dec}}^T;
    M_{2} = G_{\text{enc}} G_{\text{t}}^T ;
    M_{3} = G_{\text{dec}} G_{\text{t}}^T.
\end{gather}
The fusion results in the feature sequence \(M \in \{M_1, M_2, M_3\}\), which is used to compute the GSME loss, incorporating both adversarial and MSE losses:
\begin{align}
    \mathcal{L}_{\text{GSME}} &= \epsilon \mathcal{L}_{\text{adv}} + \frac{\theta}{4n^2} \sum_{i=1}^{3} \sum_{j}^{n}    \Big( M_{i,j}^{f} - M_{i,j}^{m} \Big)^2.
\end{align}

% By minimizing the GSME loss, our approach aligns features across the encoder, decoder, and transformer block while leveraging adversarial learning. This ensures the preservation of both high-level structural information and low-level texture details between the missing and complete modalities. As a result, the proposed method significantly enhances the network's robustness and accuracy.

\subsection{Total Loss}
The total loss function utilized during training integrates four key components, each contributing to different aspects of the model's optimization:
\begin{equation}
    \mathcal{L}_{\text{joint}} = \lambda_1 \mathcal{L}_{\text{MS-TKD}} + \lambda_2 \mathcal{L}_{\text{logit}} + \lambda_3 \mathcal{L}_{\text{GSME}} + \lambda_4 \mathcal{L}_{\text{Dice}},
\end{equation}
where \(\lambda_1, \lambda_2, \lambda_3,\) and \(\lambda_4\) are weighting coefficients that balance the contributions of each individual loss term to the overall objective.

\begin{figure}[h!]
    \centering
    \includegraphics[width=0.98\textwidth]{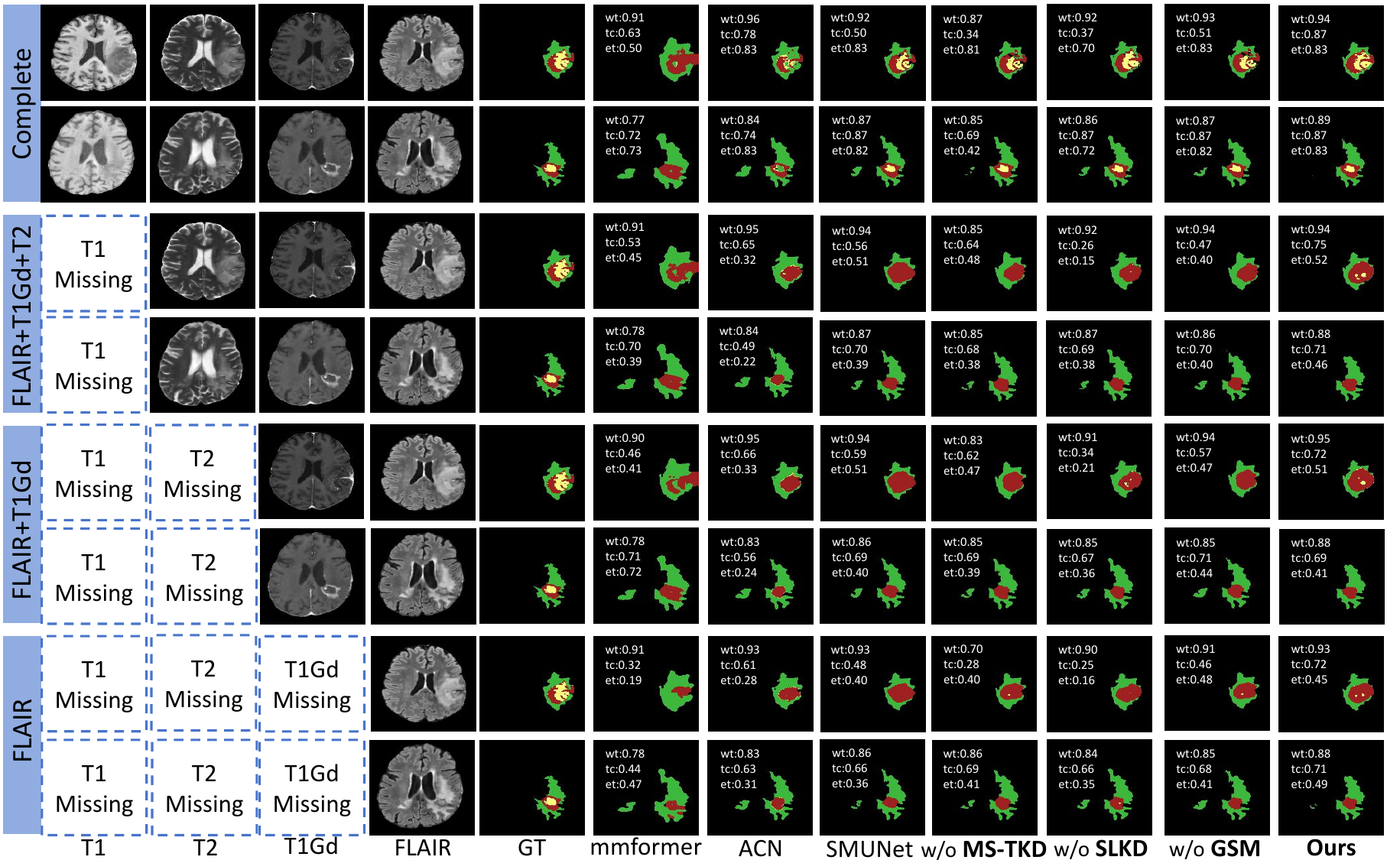}
    % \caption{Comparison of segmentation results under four missing-modality scenarios. From left to right, the figure shows T1, T2, T1ce, and FLAIR images; ground-truth labels for two patients; three columns of comparison-study results; three columns of ablation-study results; and our final segmentation. Color legend: WT = red + yellow + green, TC = red + yellow, ET = red.}
    \caption{Comparison of segmentation results under four missing-modality scenarios. Color legend: WT = \textcolor[HTML]{9F2020}{red} + \textcolor{yellow}{yellow} + \textcolor[HTML]{3DB83D}{green}, TC = \textcolor[HTML]{9F2020}{red} + \textcolor{yellow}{yellow}, ET = \textcolor[HTML]{9F2020}{red}.}
    \label{fig:2}
\end{figure}

\section{Experiments}
\subsection{Dataset and Implementation}
\textbf{BraTS 2024 dataset \cite{c21de20242024} and FeTS 2024 dataset \cite{c22pati2021federated}.}
Both datasets offer 3D multimodal MR brain images accompanied by corresponding accurate labels. The BraTS 2024 comprises 1,350 cases, and the FeTS 2024 comprises 1,250 cases, with each case including MR images in four modalities: T1, T2, T1Gd, and FLAIR. The images are classified into three distinct regions: Enhancing Tumor (ET) region, Core Tumor (CT) region, and Whole Tumor (WT) region. We randomly divided the dataset into two parts for our analysis, using 80\% for training and the remaining 20\% for testing. We applied data augmentation techniques, such as random flips, rotations, and cropping and each MRI was resized to dimensions of 160 × 192 × 128 to standardize the data.

\textbf{Implementation details.}
All experiments were conducted on a Tesla V100 GPU using PyTorch 2.4.1 as the foundational framework. The batch size for training was set to 1, and the model was trained for 250 epochs. Model parameters were optimized using the Adam optimization algorithm, with an initial learning rate of 0.0001. The hyperparameters \(\beta_1\) and \(\beta_2\) were configured to 0.9 and 0.99, respectively, to ensure stable and efficient convergence.

\subsection{Experiment Results}
\textbf{Comparative Experiments.} As shown in Tables \ref{tab:1} and \ref{tab:2}, we performed multimodal ablation experiments on the BraTS 2024 dataset. Input modes were randomly excluded to simulate various scenarios, resulting in unimodal, bimodal, trimodal, and quaternary input, for a total of 16 combinations. Compared with several state-of-the-art models, MST-KDNet achieved optimal or suboptimal results in key tumor regions (WT, TC, and ET), indicating superior voxel overlap and spatial distance performance. Notably, MST-KDNet maintains robust performance even under extreme conditions, such as when only one or two modalities were available, showing strong robustness. MST-KDNet also performs impressively on the FeTS 2024 dataset, as shown in Table \ref{tab:3}. MST-KDNet consistently outperforms the other methods in most cases of missing modal combinations.
\begin{table}[h!]
    \centering
    \fontsize{8pt}{10pt}\selectfont
    \scriptsize
    \renewcommand{\arraystretch}{1.05}
         \caption{Comparison of Dice score for various state-of-the-art models (BraTS 2024). \textcolor{red}{Red} represents the optimal value, and \textcolor{blue}{Blue} represents the suboptimal value.}
    \begin{tabularx}{\textwidth}{c|c|*{15}{Y}|Yc}  % X 自动均分列宽
        \toprule
        \multirow{4}{*}{Type} & \multirow{4}{*}{
        \begin{tabular}{c}FLAIR \\ T1 \\ T1Gd \\ T2\end{tabular}} 
        & \scalebox{2}{$\circ$} & \scalebox{2}{$\circ$} & \scalebox{2}{$\circ$} & \scalebox{2}{$\bullet$} 
        & \scalebox{2}{$\circ$} & \scalebox{2}{$\circ$} &  \scalebox{2}{$\bullet$} & \scalebox{2}{$\circ$} & \scalebox{2}{$\bullet$} & \scalebox{2}{$\bullet$} & \scalebox{2}{$\bullet$} & \scalebox{2}{$\bullet$} & \scalebox{2}{$\bullet$} & \scalebox{2}{$\circ$} & \scalebox{2}{$\bullet$} & \multirow{4}{*}{Avg.} \\
        
        &  & \scalebox{2}{$\circ$} & \scalebox{2}{$\circ$} & \scalebox{2}{$\bullet$} & \scalebox{2}{$\circ$} 
        & \scalebox{2}{$\circ$} & \scalebox{2}{$\bullet$} & \scalebox{2}{$\bullet$} & \scalebox{2}{$\bullet$} 
        & \scalebox{2}{$\circ$} & \scalebox{2}{$\circ$} & \scalebox{2}{$\bullet$} & \scalebox{2}{$\bullet$} 
        & \scalebox{2}{$\circ$} & \scalebox{2}{$\bullet$} & \scalebox{2}{$\bullet$} & \\
        
        &  & \scalebox{2}{$\circ$} & \scalebox{2}{$\bullet$} & \scalebox{2}{$\circ$} & \scalebox{2}{$\circ$} & \scalebox{2}{$\bullet$} & \scalebox{2}{$\bullet$} & \scalebox{2}{$\circ$} & \scalebox{2}{$\circ$} & \scalebox{2}{$\circ$} & \scalebox{2}{$\bullet$} & \scalebox{2}{$\bullet$} & \scalebox{2}{$\circ$} & \scalebox{2}{$\bullet$} & \scalebox{2}{$\bullet$} & \scalebox{2}{$\bullet$} & \\
        
        &  & \scalebox{2}{$\bullet$} & \scalebox{2}{$\circ$} & \scalebox{2}{$\circ$} & \scalebox{2}{$\circ$} 
        & \scalebox{2}{$\bullet$} & \scalebox{2}{$\circ$} & \scalebox{2}{$\circ$} & \scalebox{2}{$\bullet$} 
        & \scalebox{2}{$\bullet$} & \scalebox{2}{$\circ$} & \scalebox{2}{$\circ$} & \scalebox{2}{$\bullet$} 
        & \scalebox{2}{$\bullet$} & \scalebox{2}{$\bullet$} & \scalebox{2}{$\bullet$} & \\
        % \vspace{-0.2cm} % 向上调整间距
        \midrule
        
        \multirow{7}{*}{WT} 
        & RA-HVED \cite{c12jeong2022region} & \textcolor{blue}{75.4} & 51.3 & 9.5 & 71.4 & \textcolor{blue}{77.5} & 53.4 & 72.9 & 76.1 & 80.1 & 72.9 & 72.9 & 80.6 & 80.4 & \textcolor{blue}{77.7} & 80.1 & 68.8 &\\
         & RMBTS \cite{c13chen2019robust} & 70.1 & 51.2 & 51.8 & 65.0 & 75.3 & 60.6 & 76.4 & 75.0 & 77.3 & 76.0 & 79.3 & 79.7 & 80.3 & 76.1 & 80.9 & 71.7 &\\
         & mmformer \cite{c14zhang2022mmformer} & 72.6 & 55.5 & 61.3 & 72.7 & 74.3 & 65.4 & 79.2 & 75.1 & 79.6 & 78.3 & 80.0& 80.7 & 81.0 & 75.6 & 81.3 &  74.2 &\\
         & M2FTrans \cite{c15shi2023m} & 72.5 & 58.8 & 62.0 & 73.0 & 73.9 & 64.2 & 77.4 & 73.6 & 78.9 & 77.0 & 77.3 & 78.5 & 79.5 & 74.2 & 78.8 & 73.3 & \\
          & ACN \cite{c16wang2021acn} & 69.6 & 58.7 & 60.1 & 80.7 & 71.8 & 63.6 & 82.1 & 72.2 & 82.3 & 81.3 & 82.0 & 82.8 & 82.0 & 72.5 & 82.5 & 75.0 &\\
       & SMUNet \cite{c17azad2022smu} & 75.0 & \textcolor{blue}{67.9} & \textcolor{blue}{69.6} &  \textcolor{blue}{84.2} & 76.7 & \textcolor{blue}{70.6} & \textcolor{blue}{84.6} & \textcolor{blue}{77.1} & \textcolor{blue}{85.2} & \textcolor{blue}{85.2} & \textcolor{blue}{85.4} & \textcolor{blue}{85.6} & \textcolor{blue}{86.0} & 77.2 & \textcolor{blue}{86.0} & \textcolor{blue}{79.7} &\\
        & MST-KDNet & \textcolor{red}{77.2} &  \textcolor{red}{72.9} &  \textcolor{red}{73.5} &  \textcolor{red}{84.7} &  \textcolor{red}{79.8} &  \textcolor{red}{75.1} &  \textcolor{red}{85.7} &  \textcolor{red}{79.3} &  \textcolor{red}{85.8} &  \textcolor{red}{86.4} &  \textcolor{red}{86.5} &  \textcolor{red}{86.1} &  \textcolor{red}{86.9} &  \textcolor{red}{80.0} &  \textcolor{red}{86.8} &  \textcolor{red}{81.8} &\\
        \midrule
        \multirow{7}{*}{TC} 
        &  RA-HVED \cite{c12jeong2022region} & 26.5 & 54.2 & 9.4 & \textcolor{blue}{41.1} & 61.3 & 54.8 & 41.9 & 29.2 & 40.5 &  61.9 &62.5 & 43.2 & 64.0 & 61.9 & 65.0 & 47.8 & \\
        & RMBTS \cite{c13chen2019robust} & 10.9 & 36.5 & 12.6 & 11.2 & 40.4 & 37.6 & 16.8 & 15.2 & 14.5 & 38.9 & 40.1 & 17.4 & 40.4 &  40.9 & 40.6 & 27.6 &\\
         &  mmformer \cite{c14zhang2022mmformer} &  \textcolor{blue}{47.2} & 52.3 & \textcolor{blue}{44.4} & 33.1 & 62.6 & 60.6 & \textcolor{blue}{49.6} &  \textcolor{red}{51.1} & \textcolor{blue}{49.6} & 60.6 & 64.3 & \textcolor{blue}{52.6} & 65.5 & 65.3 & 67.0 & \textcolor{blue}{55.1} &\\
         & M2FTrans \cite{c15shi2023m} & 46.6 & 53.3 & 43.3 & 33.8 & 60.0 & 57.7 & 46.7 & \textcolor{blue}{48.5} & 48.3 & 57.8 & 60.0 & 49.6 & 61.5 & 60.8 & 62.0 & 52.7 & \\
         & ACN \cite{c16wang2021acn} & 21.2 & 54.2 & 19.5 & 22.5 & 58.8 & 57.9 & 26.1 & 23.2 & 26.7 & 60.0 & 63.8 & 28.3 & 62.6 & 62.7 & 64.1 &  43.4 &\\
        & SMUNet \cite{c17azad2022smu} & 29.3 & \textcolor{blue}{64.1} & 28.2 & 28.8 & \textcolor{blue}{67.3} & \textcolor{blue}{67.1} & 32.6 & 31.5 & 32.5 & \textcolor{blue}{66.9} & \textcolor{blue}{70.4} & 33.7 & \textcolor{blue}{69.4} & \textcolor{blue}{69.1} & \textcolor{blue}{69.8} & 50.7 &\\
        & MST-KDNet &  \textcolor{red}{47.3} &  \textcolor{red}{68.3} &  \textcolor{red}{44.5} &  \textcolor{red}{33.9} &  \textcolor{red}{70.3} &  \textcolor{red}{71.3} &  \textcolor{red}{50.1} & 41.5 &  \textcolor{red}{50.2} &  \textcolor{red}{72.0} &  \textcolor{red}{74.1} &  \textcolor{red}{53.6} &  \textcolor{red}{72.5} &  \textcolor{red}{72.6} &  \textcolor{red}{73.1} &  \textcolor{red}{59.5} &\\
        \midrule
        \multirow{7}{*}{ET} 
         & RA-HVED \cite{c12jeong2022region} & 35.8 & 37.8 & 9.24 & \textcolor{blue}{39.8} & 42.3 & 36.6 & 42.6 & 43.8 & 44.4 & 44.1 & 43.9 & 48.4 & 46.8 & 40.7 & 45.9 & 40.1 &\\
         & RMBTS \cite{c13chen2019robust} & 7.9 & 37.8 & 10.0 & 8.2 & 41.9 & 40.1 & 13.1 & 11.8 & 10.8 & 40.6 & 43.5 & 14.0 & 42.3 & 44.1 & 55.2 & 28.1 &\\
         & mmformer \cite{c14zhang2022mmformer} & 44.9 & 50.5 & \textcolor{blue}{42.3} & 31.4 & 61.3 & 59.0 & 45.3 & \textcolor{blue}{49.4} & 46.6 & 59.3 & 63.0 & 49.6 & 63.6 & 64.2 & 65.7 &  53.1 &\\
         & M2FTrans \cite{c15shi2023m} & \textcolor{blue}{47.1} & 54.2 &  \textcolor{red}{44.6} & 34.0 & 62.6 & 60.0 & \textcolor{blue}{47.5} & \textcolor{blue}{49.4}& \textcolor{blue}{49.3} & 60.2 & 62.7 & \textcolor{blue}{50.4} & 64.5 & 63.4 & 65.0 & \textcolor{blue}{54.3} & \\
          & ACN \cite{c16wang2021acn} & 18.0 & 55.2 & 16.9 & 19.6 & 59.8 & 59.6 & 22.2 & 19.2 & 22.4 & 60.8 & 65.1 & 23.9 & 64.0 & 64.3 & 65.9 & 42.5 &\\
        & SMUNet \cite{c17azad2022smu} & 25.5 & \textcolor{blue}{64.8} & 25.0 & 25.1 & \textcolor{blue}{67.9} & \textcolor{blue}{68.1} & 28.6 & 27.6 & 28.6 & \textcolor{blue}{67.9} & \textcolor{blue}{70.6} & 29.7 & \textcolor{blue}{69.8} & \textcolor{blue}{70.1} & \textcolor{blue}{70.8} & 49.3 &\\
        & MST-KDNet &  \textcolor{red}{48.3} &  \textcolor{red}{68.6} &  32.0 &  \textcolor{red}{40.6} &  \textcolor{red}{70.0} &  \textcolor{red}{72.3} &  \textcolor{red}{48.5} &  \textcolor{red}{50.1} &  \textcolor{red}{51.1} &  \textcolor{red}{72.4}&  \textcolor{red}{74.9} &  \textcolor{red}{52.5} &  \textcolor{red}{72.8} &   \textcolor{red}{73.1} &  \textcolor{red}{73.9} &  \textcolor{red}{59.8} &\\

        \bottomrule
    \end{tabularx}
    \label{tab:1}
\end{table}

\begin{table}[h!]
    \centering
    \fontsize{8pt}{10pt}\selectfont
    \scriptsize
    \renewcommand{\arraystretch}{1.05}
    \caption{Comparison of HD95 score for various state-of-the-art models (BraTS 2024). \textcolor{red}{Red} represents the optimal value, and \textcolor{blue}{Blue} represents the suboptimal value.}
    \begin{tabularx}{\textwidth}{c|c|*{15}{Y}|Yc}  % X 自动均分列宽
        \toprule
        \multirow{4}{*}{Type} & \multirow{4}{*}{\begin{tabular}{@{}c@{}}FLAIR \\ T1 \\ T1Gd \\ T2\end{tabular}} 
        & \scalebox{2}{$\circ$} & \scalebox{2}{$\circ$} & \scalebox{2}{$\circ$} & \scalebox{2}{$\bullet$} 
        & \scalebox{2}{$\circ$} & \scalebox{2}{$\circ$} &  \scalebox{2}{$\bullet$} & \scalebox{2}{$\circ$} & \scalebox{2}{$\bullet$} & \scalebox{2}{$\bullet$} & \scalebox{2}{$\bullet$} & \scalebox{2}{$\bullet$} & \scalebox{2}{$\bullet$} & \scalebox{2}{$\circ$} & \scalebox{2}{$\bullet$} & \multirow{4}{*}{Avg.} \\
        
        &  & \scalebox{2}{$\circ$} & \scalebox{2}{$\circ$} & \scalebox{2}{$\bullet$} & \scalebox{2}{$\circ$} 
        & \scalebox{2}{$\circ$} & \scalebox{2}{$\bullet$} & \scalebox{2}{$\bullet$} & \scalebox{2}{$\bullet$} 
        & \scalebox{2}{$\circ$} & \scalebox{2}{$\circ$} & \scalebox{2}{$\bullet$} & \scalebox{2}{$\bullet$} 
        & \scalebox{2}{$\circ$} & \scalebox{2}{$\bullet$} & \scalebox{2}{$\bullet$} & \\
        
        &  & \scalebox{2}{$\circ$} & \scalebox{2}{$\bullet$} & \scalebox{2}{$\circ$} & \scalebox{2}{$\circ$} & \scalebox{2}{$\bullet$} & \scalebox{2}{$\bullet$} & \scalebox{2}{$\circ$} & \scalebox{2}{$\circ$} & \scalebox{2}{$\circ$} & \scalebox{2}{$\bullet$} & \scalebox{2}{$\bullet$} & \scalebox{2}{$\circ$} & \scalebox{2}{$\bullet$} & \scalebox{2}{$\bullet$} & \scalebox{2}{$\bullet$} & \\
        
        &  & \scalebox{2}{$\bullet$} & \scalebox{2}{$\circ$} & \scalebox{2}{$\circ$} & \scalebox{2}{$\circ$} 
        & \scalebox{2}{$\bullet$} & \scalebox{2}{$\circ$} & \scalebox{2}{$\circ$} & \scalebox{2}{$\bullet$} 
        & \scalebox{2}{$\bullet$} & \scalebox{2}{$\circ$} & \scalebox{2}{$\circ$} & \scalebox{2}{$\bullet$} 
        & \scalebox{2}{$\bullet$} & \scalebox{2}{$\bullet$} & \scalebox{2}{$\bullet$} & \\
        \midrule
        \multirow{7}{*}{WT} 
        & RA-HVED \cite{c12jeong2022region} & 22.1 & 40.2 & 57.7 & 23.8 & 19.8 & 34.8 & 20.9 & 17.4 & 16.9 & 21.2 & 20.5 & 15.0 & 16.3 & 18.6 & 15.9 & 24.1 &\\
        & RMBTS \cite{c13chen2019robust} & 39.1 & 63.6 & 57.7 & 59.4 & 36.1 & 50.1 & 41.7 & 33.1 & 37.4 & 47.8 & 34.8 & 33.2 & 35.3 & 34.1 & 34.0 & 42.5 &\\
        & mmformer \cite{c14zhang2022mmformer} & 19.5 & 52.0 & 40.7 & 18.2 & 18.8 & 34.5 & 13.9 & 16.8 & 13.1 & 15.5 & 13.4 & 12.9 & 12.2 & 16.8 & 11.8 & 20.7 &\\
      & M2FTrans \cite{c15shi2023m} & 43.8 & 51.8 & 47.0 & 47.3 & 42.4 & 44.5 & 43.0 & 42.6 & 42.1 & 41.9 & 41.3 & 41.3 & 40.7 & 40.8 & 40.5 & 43.4 & \\
        & ACN \cite{c16wang2021acn}&11.6 & 28.4 & 29.6 & \textcolor{blue}{11.8}  & 13.5 & 20.4 & 11.4 & 15.6 &10.3 & 13.2 & 11.7 & 10.2 & 11.5 & 15.1 &10.3 & 15.0 &\\
        & SMUNet \cite{c17azad2022smu} &\textcolor{blue}{ 9.1} & \textcolor{blue}{13.3} & \textcolor{red}{5.9} & 12.2 & \textcolor{blue}{5.9} & \textcolor{red}{7.6} & \textcolor{blue}{11.2} & \textcolor{blue}{5.4} & \textcolor{blue}{7.7} & \textcolor{red}{5.1} & \textcolor{blue}{5.2} &  \textcolor{blue}{5.3} & \textcolor{blue}{4.9} & \textcolor{blue}{4.8} & \textcolor{blue}{8.0} & \textcolor{blue}{7.4} &\\
        & MST-KDNet & \textcolor{red}{8.1} & \textcolor{red}{11.1} & \textcolor{blue}{11.0} & \textcolor{red}{6.7} & \textcolor{red}{5.3} & \textcolor{blue}{9.2} & \textcolor{red}{6.1} & \textcolor{red}{5.2} & \textcolor{red}{4.6} & \textcolor{blue}{6.2} & \textcolor{red}{5.1} & \textcolor{red}{5.0} & \textcolor{red}{4.7} & \textcolor{red}{4.7} & \textcolor{red}{5.3} & \textcolor{red}{6.6} &\\
        \midrule
        \multirow{7}{*}{TC} 
        & RA-HVED \cite{c12jeong2022region} & 25.3 & 30.4 & 57.1 & 22.5 & 15.8 & 26.8 & 20.9 & 23.1 & 19.7 & 15.9 & 14.4 & 21.6 & 13.3 & 16.2 & 12.5 & 22.4 & \\
      & RMBTS \cite{c13chen2019robust} & 24.8 & 23.1 & 47.1 & 24.1 & 19.8 & 25.8 & 23.7 & 21.9 & 19.1 & 18.5 & 16.3 & 20.0 & 15.6 & 14.0 & 13.7 & 21.8 &\\
      & mmformer \cite{c14zhang2022mmformer} & 27.7 & 62.1 & 39.1 & 24.3 & 25.6 & 38.7 & 19.7 & 24.1 & 19.3 & 20.5 & 17.3 & 18.7 & 15.4 & 22.1 & 14.7 & 26.0 &\\
      & M2FTrans \cite{c15shi2023m} & 79.4 & 79.2 & 82.6 & 82.4 & 76.3 & 76.3 & 79.7 & 79.2 & 79.5 & 78.5 & 77.5 & 78.3 & 77.0 & 77.0 & 76.3 & 78.6 & \\
        & ACN \cite{c16wang2021acn} & 15.7 & 9.2 & 19.3 & 18.2 & 6.4 & 8.5 & 17.3 & 17.0 & 15.7 & 6.6 & 6.2 & 17.6 & 5.8 & 6.2 & 5.8 & 11.7 &\\
        & SMUNet \cite{c17azad2022smu} & \textcolor{blue}{14.0} & \textcolor{blue}{6.3} & \textcolor{blue}{14.0} & \textcolor{blue}{13.4} & \textcolor{blue}{4.4} & \textcolor{blue}{5.0} & \textcolor{blue}{12.2} & \textcolor{blue}{12.1} & \textcolor{blue}{12.0} & \textcolor{blue}{4.8} & \textcolor{blue}{4.3 }& \textcolor{blue}{11.9} & \textcolor{blue}{4.2} & \textcolor{blue}{4.5} & \textcolor{blue}{4.6} & \textcolor{blue}{8.5} &\\  
      & MST-KDNet & \textcolor{red}{12.0} & \textcolor{red}{4.9} & \textcolor{red}{11.2} & \textcolor{red}{12.1} & \textcolor{red}{3.7} & \textcolor{red}{4.3} & \textcolor{red}{10.5} & \textcolor{red}{10.8} & \textcolor{red}{11.0} & \textcolor{red}{3.6} & \textcolor{red}{3.4} & \textcolor{red}{10.0} & \textcolor{red}{3.7} & \textcolor{red}{3.3} & \textcolor{red}{4.0} & \textcolor{red}{7.2} &\\
        \midrule
        \multirow{7}{*}{ET} 
      & RA-HVED \cite{c12jeong2022region}	 & \textcolor{blue}{12.9} & 25.0 & 47.0 & 15.2 & 14.9 & 23.7 & 13.2 & \textcolor{blue}{10.9} & \textcolor{blue}{10.8} & 14.0 & 14.2 & \textcolor{blue}{11.0} & 12.8 & 15.4 & 12.2 & 16.9 &\\
      & RMBTS \cite{c13chen2019robust} & 23.8 & 21.9 & 44.8 & 23.7 & 19.2 & 24.2 & 22.4 & 21.9 & 19.5 & 17.2 & 15.1 & 19.5 & 15.2 & 13.5 & 13.3 & 21.0 &\\ 
      & mmformer \cite{c14zhang2022mmformer} & 26.4 & 59.8 & 37.6 & 23.2 & 24.0 & 36.7 & 18.6 & 22.2 & 18.4 & 18.3 & 16.4 & 17.7 & 14.5 & 20.4 & 14.0 & 24.5 &\\ 
      & M2FTrans \cite{c15shi2023m} & 23.4 & 31.5 & 21.5 & 24.1 & 16.1 & 16.2 & 16.2 & 19.4 & 20.9 & 16.8 & 13.3 & 18.5 & 15.3 & 14.2 & 13.9 & 18.8 &  \\
      & ACN \cite{c16wang2021acn} & 14.7 & 8.0 & 19.3 & 18.1 & 6.1 & 7.6 & 16.6 & 16.4 & 14.9 & 5.9 & 5.3 & 17.2 & 5.2 & 5.3 & 5.2  & 11.1 &\\
        & SMUNet \cite{c17azad2022smu} & 13.5 & \textcolor{blue}{5.4} & \textcolor{blue}{14.0} & \textcolor{blue}{13.0} & \textcolor{blue}{3.9 }& \textcolor{blue}{4.3} & \textcolor{blue}{11.8} & 11.5 & 12.0 & \textcolor{blue}{4.1} & \textcolor{blue}{3.7}& 11.3 & \textcolor{blue}{3.7} & \textcolor{blue}{4.0} & \textcolor{blue}{4.0 }& \textcolor{blue}{8.0} & \\ 
      & MST-KDNet & \textcolor{red}{11.7} & \textcolor{red}{4.5} & \textcolor{red}{10.5} & \textcolor{red}{11.9} & \textcolor{red}{3.3 }& \textcolor{red}{3.8} & \textcolor{red}{9.8 }& \textcolor{red}{10.3} & \textcolor{red}{10.6} & \textcolor{red}{3.2} & \textcolor{red}{3.0} & \textcolor{red}{9.8} & \textcolor{red}{3.3} & \textcolor{red}{2.9} & \textcolor{red}{3.0} & \textcolor{red}{6.8} &\\
        \bottomrule
    \end{tabularx}
    \label{tab:2}
\end{table}

\textbf{Ablation Study.} As shown in Table \ref{tab:5}, on the BraTS 2024 dataset, excluding MS-TKD decreased Dice scores by 2.0\% for WT, 5.1\% for TC, and 5.6\% for ET, while increasing HD95, highlighting the importance of multiscale attentional alignment. Removing GSME reduced Dice scores by 3.5\% for WT, 3.4\% for TC, and 6.4\% for ET, emphasizing the role of global style and texture compensation. The absence of SLKD caused Dice score drops of 1.8\% for WT, 3.4\% for TC, and 4.6\% for ET, indicating the importance of flexible teacher-student distribution matching. A similar pattern was observed in the FeTS 2024 dataset. These results indicate that MS-TKD, GSME, and SLKD have their respective and complementary roles, which significantly improve the model's segmentation accuracy and stability under missing mode conditions.

\begin{table}[h!]
\centering
\fontsize{8pt}{10pt}\selectfont
\caption{Comparison of average Dice and HD95 scores for various state-of-the-art models (FeTS 2024). \textcolor{orange}{Due to space limits, full 16 results are in our \href{https://github.com/Quanato607/MST-KDNet}{GitHub}}.}
\label{tab:3}
    \centering
    
    \begin{adjustbox}{width=1\textwidth, center}
\begin{tabularx}{\textwidth}{c|>{\centering\arraybackslash}X|>{\centering\arraybackslash}X|>{\centering\arraybackslash}X|>{\centering\arraybackslash}X|>{\centering\arraybackslash}X|>{\centering\arraybackslash}X}  % 修改对齐方式
    % \hline
    % \multicolumn{7}{c}{\centering FeTS 2024 \cite{c21de20242024}} \\
    \hline
    \multirow{2}*{\centering \textbf{Method}} & \multicolumn{3}{c|}{\textbf{Average Dice Score (\%) }} & \multicolumn{3}{c}{\textbf{Average HD95 Score (mm)}} \\
    \cline{2-7}
    ~ & \centering WT & \centering TC & \centering ET & \centering WT & \centering TC & ET \\
    \hline
    RA-HVED \cite{c12jeong2022region} & 69.7 & 60.0 & 50.9 & 22.0 & 20.6 & 19.8 \\
    \hline
    RMBTS \cite{c13chen2019robust} & 75.2 & 60.4 & 65.6 & 8.6 & 25.2 & 19.1 \\
    \hline
    mmformer \cite{c14zhang2022mmformer} & 68.9 & 54.6& 48.6 & 26.7 &27.5 & 34.0 \\
    \hline
    M2FTrans \cite{c15shi2023m} & 82.0 & 74.3& 63.0 & 26.5 &14.8 & 20.8\\
    \hline
    ACN \cite{c16wang2021acn} & 84.9 & 78.8& 67.3 & 8.5 &8.4 & 16.5\\
    \hline
    SMUNet \cite{c17azad2022smu} & \textcolor{blue}{87.5} & \textcolor{blue}{82.9}& \textcolor{blue}{72.1} & \textcolor{blue}{6.4} &\textcolor{blue}{6.3} & \textcolor{blue}{5.5}\\
    \hline
    MST-KDNet & \textcolor{red}{88.4} & \textcolor{red}{84.3} & \textcolor{red}{73.4} & \textcolor{red}{5.9} & \textcolor{red}{5.7} & \textcolor{red}{5.4} \\
    \hline
\end{tabularx}
    \end{adjustbox}
\end{table}

\begin{table}[h!]
\centering
\fontsize{8pt}{10pt}\selectfont
\caption{Comparison of average Dice and HD95 scores for ablation studys.}
\label{tab:5}
    \centering
    \begin{adjustbox}{width=1\textwidth, center}
\begin{tabularx}{\textwidth}{c|>{\centering\arraybackslash}X|>{\centering\arraybackslash}X|>{\centering\arraybackslash}X|>{\centering\arraybackslash}X|>{\centering\arraybackslash}X|>{\centering\arraybackslash}X}  % 修改对齐方式
    \hline
    \multicolumn{7}{c}{\centering \textbf{BraTS 2024} \cite{c21de20242024}} \\
    \hline
    \multirow{2}*{\centering \textbf{Method}} & \multicolumn{3}{c|}{\textbf{Average Dice Score (\%) }} & \multicolumn{3}{c}{\textbf{Average HD95 Score (mm)}} \\
    \cline{2-7}
    ~ & \centering WT & \centering TC & \centering ET & \centering WT & \centering TC & ET \\
    \hline
    w/o MS-TKD & 79.8 & 54.4 & 54.2 & \textcolor{blue}{7.5} & \textcolor{blue}{8.3} & \textcolor{blue}{7.8} \\
    \hline
    w/o GSME & 78.3 & 55.1 & 53.4 & 9.6 & 9.7 & 9.5 \\
    \hline
    w/o SLKD & \textcolor{blue}{80.0} & \textcolor{blue}{56.1} & \textcolor{blue}{55.2} & 8.1 & 8.7 & 8.0 \\
    \hline
    MST-KDNet & \textcolor{red}{81.8} & \textcolor{red}{59.5} & \textcolor{red}{59.8} & \textcolor{red}{6.6} & \textcolor{red}{7.2} & \textcolor{red}{6.8} \\
    \hline
    \multicolumn{7}{c}{\centering \textbf{FeTS 2024} \cite{c22pati2021federated}} \\
    \hline
    % \multirow{2}*{\centering Method} & \multicolumn{3}{c|}{\textbf{Average Dice Score (\%) }} & \multicolumn{3}{c}{\textbf{Average HD95 Score (mm)}} \\
    % \cline{2-7}
    % ~ & \centering WT & \centering TC & \centering ET & \centering WT & \centering TC & ET \\
    % \hline
    w/o MS-TKD & 87.0 & 81.8 & 72.6 & 7.3 & 6.8 & 5.5 \\
    \hline
    w/o GSME & 86.1 & \textcolor{blue}{82.9} & 72.6 & 7.3 & 6.6 & 5.9 \\
    \hline
    w/o SLKD & \textcolor{blue}{87.5} & 82.1 & \textcolor{blue}{72.9} & \textcolor{blue}{6.5} & \textcolor{blue}{6.6} & \textcolor{blue}{5.8} \\
    \hline
    MST-KDNet & \textcolor{red}{88.2} & \textcolor{red}{84.3} & \textcolor{red}{73.4} & \textcolor{red}{5.9} & \textcolor{red}{5.7} & \textcolor{red}{5.4} \\
    \hline
\end{tabularx}
    \end{adjustbox}
\end{table}

% In this work, we introduce MST-KDNet, a novel framework designed for incomplete multi-modality brain tumor segmentation. MST-KDNet effectively captures cross-modality correlations while enhancing tumor region representations, enabling robust and accurate segmentation even in scenarios with significant modality loss. To address redundancy in modality-specific features, the framework employs global and local feature refinement mechanisms. These refinements align the available modalities while compensating for the missing ones, resulting in a more balanced feature distribution across modalities. This alignment strengthens the model's overall segmentation performance and robustness. Extensive experiments conducted on the BraTS 2024 and FeTS 2024 benchmarks demonstrate the superiority and adaptability of MST-KDNet. At the same time, compared to state-of-the-art methods, our proposed approach consistently achieves improved performance, particularly in incomplete multi-modality settings, further affirming its effectiveness and reliability.

\section{Conclusion}
In this study, we propose MST-KDNet, a novel framework for incomplete multi-modality brain tumor segmentation. MST-KDNet effectively captures cross-modality correlations and significantly enhances tumor region representations for robust segmentation, even with significant missing modalities. The framework employs global and local feature refinement mechanisms to align available modalities, effectively compensating for the missing ones and improving feature distribution. Extensive experiments on the BraTS and FeTS 2024 benchmarks demonstrate MST-KDNet's superiority and robustness, consistently outperforming state-of-the-art methods, especially in incomplete modality settings.

\subsubsection{Acknowledgements} This work was supported by the Fundamental Research Funds for the Provincial Universities of Zhejiang (No. GK259909299001-006), Anhui Provincial Joint Construction Key Laboratory of Intelligent Education Equipment and Technology (No. IEET202401), the Guangxi Key R\&D Project (No. AB24010167), the Project (No. 20232ABC03A25), Guangdong Basic and Applied Basic Research Foundation (No. 2025A1515011617), Shenzhen Medical Research Fund (No. C2401036), and Hospital University United Fund of The Second Affiliated Hospital, School of Medicine, The Chinese University of Hong Kong, Shenzhen (No. HUUF-MS-202303). 

\subsubsection{Disclosure of Interests} The authors declare no competing interests.

\clearpage

\bibliographystyle{splncs04}
%\bibliography{Paper-2782}

\begin{thebibliography}{10}
\providecommand{\url}[1]{\texttt{#1}}
\providecommand{\urlprefix}{URL }
\providecommand{\doi}[1]{https://doi.org/#1}

\bibitem{c17azad2022smu}
Azad, R., Khosravi, N., Merhof, D.: Smu-net: Style matching u-net for brain tumor segmentation with missing modalities. In: International Conference on Medical Imaging with Deep Learning. pp. 48--62. PMLR (2022)

\bibitem{c13chen2019robust}
Chen, C., Dou, Q., Jin, Y., Chen, H., Qin, J., Heng, P.A.: Robust multimodal brain tumor segmentation via feature disentanglement and gated fusion. In: International Conference on Medical Image Computing and Computer-Assisted Intervention. pp. 447--456. Springer (2019)

\bibitem{c28chen2025toward}
Chen, Y., Zhu, S., Fang, Z., Liu, C., Zou, B., Qiu, L., Wang, Y., Chang, S., Jia, F., Qin, F., Fan, J., Peng, Y., Wang, C.: Toward robust early detection of alzheimer’s disease via an integrated multimodal learning approach. In: ICASSP 2025 - 2025 IEEE International Conference on Acoustics, Speech and Signal Processing (ICASSP). pp.~1--5 (2025)

\bibitem{c7dai2024federated}
Dai, Q., Wei, D., Liu, H., Sun, J., Wang, L., Zheng, Y.: Federated modality-specific encoders and multimodal anchors for personalized brain tumor segmentation. In: Proceedings of the AAAI Conference on Artificial Intelligence. vol.~38, pp. 1445--1453 (2024)

\bibitem{c4ding2021rfnet}
Ding, Y., Yu, X., Yang, Y.: Rfnet: Region-aware fusion network for incomplete multi-modal brain tumor segmentation. In: Proceedings of the IEEE/CVF International Conference on Computer Vision. pp. 3975--3984 (2021)

\bibitem{c26ge2024tc}
Ge, R., Yu, X., Chen, Y., Zhou, G., Jia, F., Zhu, S., Jia, J., Zhang, C., Sun, Y., Zeng, D., et~al.: Tc-kanrecon: High-quality and accelerated mri reconstruction via adaptive kan mechanisms and intelligent feature scaling. arXiv preprint arXiv:2408.05705  (2024)

\bibitem{c20hatamizadeh2022unetr}
Hatamizadeh, A., Tang, Y., Nath, V., Yang, D., Myronenko, A., Landman, B., Roth, H.R., Xu, D.: Unetr: Transformers for 3d medical image segmentation. In: Proceedings of the IEEE/CVF Winter Conference on Applications of Computer Vision. pp. 574--584 (2022)

\bibitem{c10huo2024c2kd}
Huo, F., Xu, W., Guo, J., Wang, H., Guo, S.: C2kd: Bridging the modality gap for cross-modal knowledge distillation. In: Proceedings of the IEEE/CVF Conference on Computer Vision and Pattern Recognition. pp. 16006--16015 (2024)

\bibitem{c12jeong2022region}
Jeong, S., Cho, H., Kwon, J., Park, H.: Region-of-interest attentive heteromodal variational encoder-decoder for segmentation with missing modalities. In: Proceedings of the Asian Conference on Computer Vision. pp. 3707--3723 (2022)

\bibitem{c9liu2023m3ae}
Liu, H., Wei, D., Lu, D., Sun, J., Wang, L., Zheng, Y.: M3ae: multimodal representation learning for brain tumor segmentation with missing modalities. In: Proceedings of the AAAI Conference on Artificial Intelligence. vol.~37, pp. 1657--1665 (2023)

\bibitem{c24liu2024anomaly}
Liu, M., Jiao, Y., Lu, J., Chen, H.: Anomaly detection for medical images using teacher-student model with skip connections and multi-scale anomaly consistency. IEEE Transactions on Instrumentation and Measurement  (2024)

\bibitem{c22pati2021federated}
Pati, S., Baid, U., Zenk, M., Edwards, B., Sheller, M., Reina, G.A., Foley, P., Gruzdev, A., Martin, J., Albarqouni, S., et~al.: The federated tumor segmentation (fets) challenge. arXiv preprint arXiv:2105.05874  (2021)

\bibitem{c15shi2023m}
Shi, J., Yu, L., Cheng, Q., Yang, X., Cheng, K.T., Yan, Z.: M2ftrans: Modality-masked fusion transformer for incomplete multi-modality brain tumor segmentation. IEEE Journal of Biomedical and Health Informatics  (2023)

\bibitem{c25sun2024logit}
Sun, S., Ren, W., Li, J., Wang, R., Cao, X.: Logit standardization in knowledge distillation. In: Proceedings of the IEEE/CVF Conference on Computer Vision and Pattern Recognition. pp. 15731--15740 (2024)

\bibitem{c3varsavsky2018pimms}
Varsavsky, T., Eaton-Rosen, Z., Sudre, C.H., Nachev, P., Cardoso, M.J.: Pimms: permutation invariant multi-modal segmentation. In: International Conference on Medical Image Computing and Computer-Assisted Intervention. pp. 201--209. Springer (2018)

\bibitem{c21de20242024}
de~Verdier, M.C., Saluja, R., Gagnon, L., LaBella, D., Baid, U., Tahon, N.H., Foltyn-Dumitru, M., Zhang, J., Alafif, M., Baig, S., et~al.: The 2024 brain tumor segmentation (brats) challenge: Glioma segmentation on post-treatment mri. arXiv preprint arXiv:2405.18368  (2024)

\bibitem{c8wang2023learnable}
Wang, H., Ma, C., Zhang, J., Zhang, Y., Avery, J., Hull, L., Carneiro, G.: Learnable cross-modal knowledge distillation for multi-modal learning with missing modality. In: International Conference on Medical Image Computing and Computer-Assisted Intervention. pp. 216--226. Springer (2023)

\bibitem{c16wang2021acn}
Wang, Y., Zhang, Y., Liu, Y., Lin, Z., Tian, J., Zhong, C., Shi, Z., Fan, J., He, Z.: Acn: adversarial co-training network for brain tumor segmentation with missing modalities. In: International Conference on Medical Image Computing and Computer-Assisted Intervention. pp. 410--420. Springer (2021)

\bibitem{c1weller2015glioma}
Weller, M., Wick, W., Aldape, K., Brada, M., Berger, M., Pfister, S.M., Nishikawa, R., Rosenthal, M., Wen, P.Y., Stupp, R., et~al.: Glioma. Nature Reviews Disease Primers  \textbf{1}(1),  1--18 (2015)

\bibitem{c27wu2025towards}
Wu, C., Chen, Y., Du, Y., Zong, J., Dong, J., Liu, M., Peng, Y., Fan, J., Qin, F., Wang, C.: Towards practical alzheimer's disease diagnosis: A lightweight and interpretable spiking neural model. arXiv preprint arXiv:2506.09695  (2025)

\bibitem{c11xing2024comprehensive}
Xing, X., Zhu, M., Chen, Z., Yuan, Y.: Comprehensive learning and adaptive teaching: Distilling multi-modal knowledge for pathological glioma grading. Medical Image Analysis  \textbf{91},  102990 (2024)

\bibitem{c19zhang2024tc}
Zhang, C., Chen, Y., Fan, Z., Huang, Y., Weng, W., Ge, R., Zeng, D., Wang, C.: Tc-diffrecon: Texture coordination mri reconstruction method based on diffusion model and modified mf-unet method. In: 2024 IEEE International Symposium on Biomedical Imaging (ISBI). pp.~1--5 (2024)

\bibitem{c6zhang2023multi}
Zhang, J., Zhang, S., Shen, X., Lukasiewicz, T., Xu, Z.: Multi-condos: Multimodal contrastive domain sharing generative adversarial networks for self-supervised medical image segmentation. IEEE Transactions on Medical Imaging  \textbf{43}(1),  76--95 (2023)

\bibitem{c14zhang2022mmformer}
Zhang, Y., He, N., Yang, J., Li, Y., Wei, D., Huang, Y., Zhang, Y., He, Z., Zheng, Y.: mmformer: Multimodal medical transformer for incomplete multimodal learning of brain tumor segmentation. In: International Conference on Medical Image Computing and Computer-Assisted Intervention. pp. 107--117. Springer (2022)

\bibitem{c5zhao2022modality}
Zhao, Z., Yang, H., Sun, J.: Modality-adaptive feature interaction for brain tumor segmentation with missing modalities. In: International Conference on Medical Image Computing and Computer-Assisted Intervention. pp. 183--192. Springer (2022)

\bibitem{c18zhu2024xlstm}
Zhu, S., Chen, Y., Jiang, S., Chen, W., Liu, C., Wang, Y., Chen, X., Ke, Y., Qin, F., Wang, C., Zhu, Z.: Xlstm-hved: Cross-modal brain tumor segmentation and mri reconstruction method using vision xlstm and heteromodal variational encoder-decoder. In: 2025 IEEE 22nd International Symposium on Biomedical Imaging (ISBI). pp.~1--5 (2025)

\bibitem{c2zhu2024brain}
Zhu, Z., Wang, Z., Qi, G., Mazur, N., Yang, P., Liu, Y.: Brain tumor segmentation in mri with multi-modality spatial information enhancement and boundary shape correction. Pattern Recognition  \textbf{153},  110553 (2024)

\end{thebibliography}

\end{document}